\documentclass[11pt]{article}

\usepackage[final]{acl}

\usepackage{times}
\usepackage{latexsym}

\usepackage[T1]{fontenc}

\usepackage[utf8]{inputenc}

\usepackage{microtype}

\usepackage{inconsolata}

\usepackage{hyperref}
\usepackage{url}
\usepackage{graphicx}
\usepackage{multirow}
\usepackage{booktabs}
\usepackage{amsmath}
\usepackage{amsfonts}
\usepackage{natbib}
\usepackage{subcaption}
\usepackage{arydshln}
\usepackage{xcolor}
\usepackage[most]{tcolorbox}

\newtcolorbox{promptbox}[1][]{%
  enhanced, breakable, boxrule=0.6pt, arc=2pt,
  colback=gray!4, colframe=gray!60,
  left=8pt, right=8pt, top=6pt, bottom=6pt,
  fonttitle=\bfseries\small,
  coltitle=black, attach boxed title to top left={yshift=-2mm,xshift=4mm},
  boxed title style={colback=white, boxrule=0.4pt, colframe=gray!60, sharp corners=all},
  before skip=4pt, after skip=4pt, #1
}

\title{MLLMCLIP: Feature-Level Distillation of MLLM for Robust Vision-Language Representations}

\author{
 \textbf{Jongsuk Kim\textsuperscript{1,2}}\thanks{Currently at PYLER. Work done during the internship at Sony Group Corporation.},
 \textbf{Qiyu Wu\textsuperscript{2}\thanks{Corresponding Author: \texttt{qiyu.wu@sony.com}. }},
 \textbf{Zhuoyuan Mao\textsuperscript{2}\thanks{Currently at Tencent. Work done while at Sony Group
Corporation.}},
 \textbf{Hiromi Wakaki\textsuperscript{2}},
\\
 \textbf{Junmo Kim\textsuperscript{1}},
 \textbf{Yuki Mitsufuji\textsuperscript{2}}
\\
 \textsuperscript{1}KAIST,
 \textsuperscript{2}Sony Group Corporation
 \\
 \texttt{js.kim@pyler.tech} ~
 \textsuperscript{2}\texttt{\{firstname.lastname\}@sony.com} ~
}

\begin{document}
\maketitle
\begin{abstract}
Pretrained vision-language models such as CLIP excel at zero-shot recognition but often fail at compositionality, particularly attribute-object and relational structures.
Recent studies mitigate this issue by augmenting training with synthetic hard negatives generated by a cascade of large language models and text-to-image models, which incurs substantial pipeline overhead.
We instead propose \textbf{MLLMCLIP}, a \emph{heterogeneous} distillation framework that transfers multimodal knowledge directly from a generative Multimodal Large Language Model~(MLLM) teacher into a discriminative CLIP student, bypassing synthetic data entirely.
To bridge the architectural mismatch between the two paradigms, we introduce an attention-based per-layer token selection and a CKA-based distillation loss.
Compared to prior CLIP-enhancement methods, MLLMCLIP achieves state-of-the-art compositional accuracy while delivering consistent gains on standard zero-shot classification and image-text retrieval, showing that feature-level distillation strengthens both compositional and general vision-language representation capability.
\end{abstract}
\section{Introduction}
\begin{figure}[t]
    \centering
    \begin{subfigure}[t]{\linewidth}
        \centering
        \includegraphics[width=\linewidth]{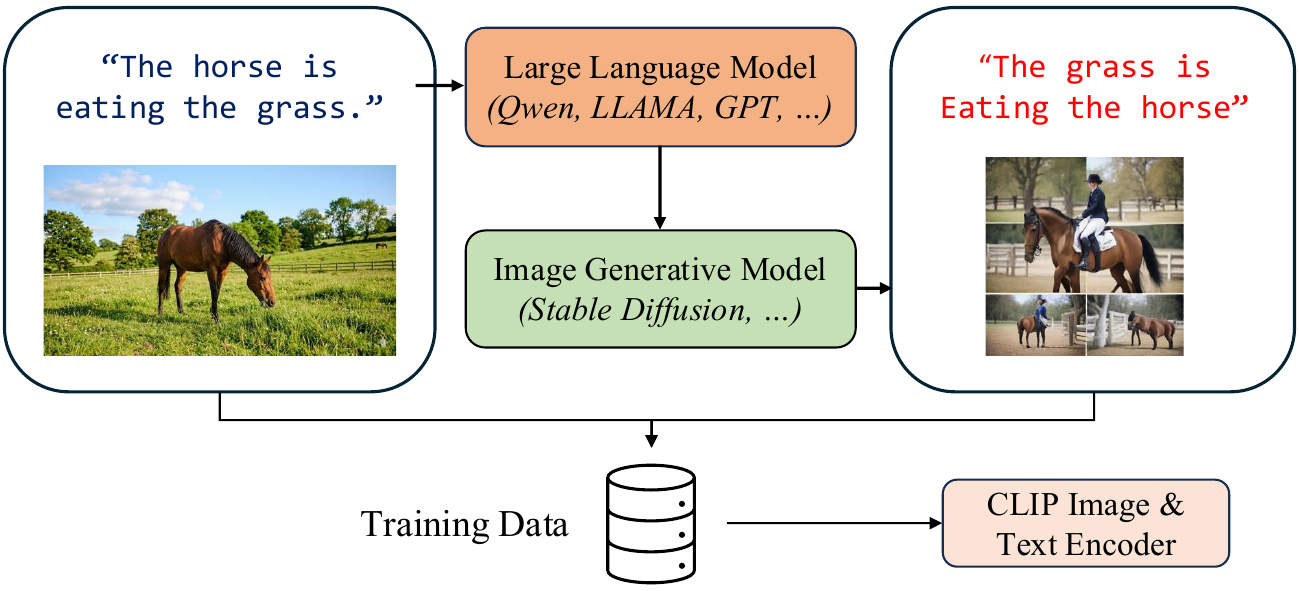}
        \caption{Data-level distillation (\textit{NegCLIP, TripletCLIP, ...})}
        \label{fig:intro_a}
    \end{subfigure}
    \vfill
    \vspace{1em}
    \begin{subfigure}[t]{\linewidth}
        \centering
        \includegraphics[width=\linewidth]{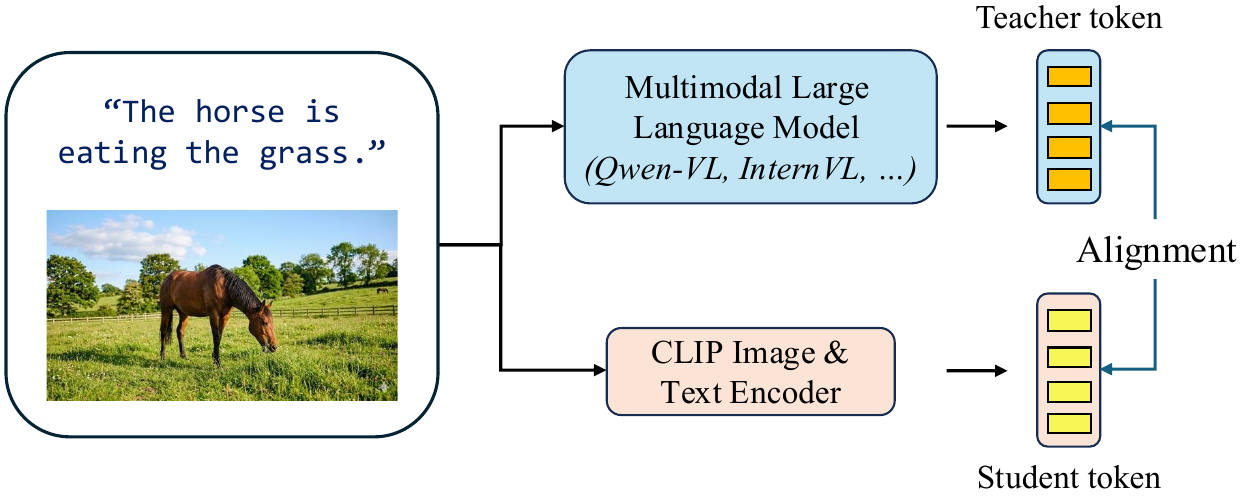}
        \caption{Feature-level distillation (Ours)}
        \label{fig:intro_b}
    \end{subfigure}
    \caption{Two distillation paradigms. (\subref{fig:intro_a})~Prior methods synthesize hard-negative samples via cascaded LLM and image generators. (\subref{fig:intro_b})~We distill hidden states from a frozen MLLM into CLIP directly.}
    \label{fig:intro}
    \vspace{-1em}
\end{figure}

The emergence of CLIP~\citep{radford2021learning} marked a turning point in vision-language learning, showing strong performance in zero-shot classification and retrieval.
Following its success, a wide range of approaches have been proposed to enhance CLIP, with a focus on embedding space~\citep{goel2022cyclip, lee2022uniclip, lavoie2024modeling}, data efficiency~\citep{li2021supervision, joshi2024data}, and higher-quality captions~\citep{fan2023improving, lai2024veclip}.
Despite these advances, a critical limitation has emerged: CLIP often exhibits a bag-of-words behavior~\citep{yuksekgonul2022and}, which prevents it from capturing the compositionality.

Constructing hard negatives that require compositional reasoning is a common direction to mitigate this issue.
Recent studies~\citep{wu2023towards, patel2024tripletclip, singh2025learning} pursue this through what we term \emph{data-level distillation}: external Large Language Models (LLM) and text-to-image generative models are used to synthesize hard-negative captions and images, and the teacher knowledge is delivered to CLIP only through these synthesized samples.
However, this paradigm suffers from two key limitations:
(1) the lack of unified multimodal understanding, as it relies on separate experts; and
(2) the inherent inefficiency of its sequential data-generation pipeline.
These shortcomings motivate a shift from data-level to \emph{feature-level distillation}: rather than expressing teacher knowledge as additional training samples, we transfer it directly through the hidden-state features of a unified Multimodal Large Language Model (MLLM).
As shown in Figure~\ref{fig:intro}, we instantiate this idea as \textbf{MLLMCLIP}, which aligns the hidden states of an MLLM teacher with a CLIP student in a single forward pass, removing the iterative sampling and quality-control overhead of synthetic-data pipelines.

Realizing this idea requires solving a problem fundamentally different from prior distillation work.
Conventional CLIP-to-CLIP~\citep{yang2024clip} and MLLM-to-MLLM~\citep{cai2025llava} distillation operate within a single architectural family.
In contrast, our setting is \emph{heterogeneous}: a generative MLLM decoder serves as the teacher and a discriminative CLIP encoder as the student.

We first conduct a pilot study showing that base MLLMs exhibit strong compositional understanding while their embedding-tuned counterparts do not, motivating distillation from the base MLLMs.
We then address two design challenges induced by heterogeneity.
To select the most informative teacher signal without layer correspondence, we propose an attention-based token selection.
To stably bridge the feature-space gap, we adopt a variant of CKA~\citep{kornblith2019similarity} loss that aligns structural relationships.

We evaluate MLLMCLIP on 11 compositionality benchmarks, 13 zero-shot classification datasets, and 2 image-text retrieval benchmarks, consistently outperforming prior CLIP-enhancement methods and achieving the highest average score across all regimes.
Moreover, MLLMCLIP achieves competitive compositional accuracy with an order-of-magnitude lower compute cost than recent MLLM-as-embedding approaches~\citep{meng2025vlm2vec}.
The contribution of our paper is summarized as follows:
\begin{itemize}
    \item We propose a heterogeneous distillation framework that transfers multimodal knowledge from a generative MLLM decoder into a discriminative CLIP encoder.
    \item We design a distillation pipeline tailored to the architectural mismatch, combining attention-based teacher-token selection with a CKA-based feature alignment loss.
    \item MLLMCLIP outperforms previous CLIP-enhancement methods on 26 benchmarks across compositionality, classification, and retrieval with diverse MLLM teachers.
\end{itemize}

\section{Related Works}
\subsection{Compositionality}
The success of CLIP has inspired numerous studies~\citep{mu2022slip, lavoie2024modeling, zheng2024dreamlip} aimed at enhancing its generalizability through data augmentation, improved training strategies, and enriched textual supervision, such as incorporating paraphrased or longer captions during training.
In contrast to these general enhancements, \citet{yuksekgonul2022and} identifies a key limitation of CLIP: its tendency to behave like a bag-of-words model.
This observation introduces the issue of compositionality, which has led to the development of new benchmarks~\citep{krojer2022image, peng2024synthesize, dumpala2024sugarcrepe++} specifically designed to assess compositional reasoning in vision-language models.
To address this problem, prior work has explored several strategies for constructing hard negative pairs for contrastive learning.
Early works employ WordNet~\citep{fellbaum2010wordnet} to generate semantically challenging negative texts~\citep{yuksekgonul2022and, oh2024preserving}.
Subsequent methods utilize LLMs to synthesize negative texts and further leverage text-to-image generative models to produce corresponding negative images, enabling sequential construction of multimodal negatives~\citep{wu2023towards, patel2024tripletclip, singh2025learning}.

\subsection{Distillation Paradigms}
We distinguish two regimes in knowledge distillation based on the architectural relationship between teacher and student.
Most prior efforts are \emph{homogeneous}: a larger model is compressed into a smaller model within the same family.
This direction has been explored extensively for CLIP-to-CLIP distillation~\citep{yang2024clip, chen2024comkd}, LLM-to-LLM distillation~\citep{chenglin2024mixed, xu2024survey}, and MLLM-to-MLLM distillation~\citep{cai2025llava, xu2024llavadi}, where the knowledge being transferred remains within the same paradigm.
In contrast, distilling from the generative decoder of an MLLM into the discriminative encoders of CLIP introduces a fundamental architectural and paradigmatic mismatch.
MLLMCLIP is a \emph{heterogeneous} distillation framework that bridges these two families, transferring the multimodal understanding of a generative MLLM into a discriminative CLIP student.

\begin{table}[t]
\centering
\small
\caption{SugarCrepe accuracy of representative MLLM-as-judge and embedding-based teachers. Embedding models are position-invariant; full results are in Table~\ref{tab:full_sugarcrepe_results}.}
\label{tab:pilot}
\setlength{\tabcolsep}{4pt}
\resizebox{\linewidth}{!}{%
\begin{tabular}{llrrr}
\toprule
Type & Model (Backbone) & First & Second & Avg. \\
\midrule
\multirow{6}{*}{MLLM}
  & LLAVA-1.6-mistral-7B   & 95.6 & 74.3 & 85.0 \\
  & LLaMA-3.2-Vision-11B   & 92.1 & 92.5 & 92.3 \\
  & Qwen2-VL-2B            & 88.9 & 95.4 & 92.2 \\
  & Qwen3-VL-2B            & 94.0 & 95.0 & 94.5 \\
  & Qwen3.5-2B             & 88.5 & 96.0 & 92.3 \\
  & InternVL3.5-2B         & 94.2 & 89.4 & 91.8 \\
\midrule
\multirow{6}{*}{Embed.}
  & VLM2Vec-v1                       & \multicolumn{2}{c}{\multirow{2}{*}{-}} & \multirow{2}{*}{66.7} \\
  & \textit{(LLAVA-1.6-mistral-7B)}       & \multicolumn{2}{c}{} & \\
  & VLM2Vec-v2                       & \multicolumn{2}{c}{\multirow{2}{*}{-}} & \multirow{2}{*}{72.7} \\
  & \textit{(Qwen2-VL-2B)}                & \multicolumn{2}{c}{} & \\
  & Qwen3-VL-Embedding-2B            & \multicolumn{2}{c}{\multirow{2}{*}{-}} & \multirow{2}{*}{83.0} \\
  & \textit{(Qwen3-VL-2B)}                & \multicolumn{2}{c}{} & \\
\bottomrule
\end{tabular}%
}
\end{table}

\section{Does MLLM Possess Sufficient Compositionality?}\label{sec:pilot}
Before constructing a distillation framework, we first ask a prerequisite question: \emph{do current MLLMs themselves understand compositional structures well enough to act as a teacher?}
A model that struggles with compositional benchmarks cannot transfer compositional knowledge to a student.
We therefore conduct a pilot study that measures the compositional understanding of recent MLLMs.

\paragraph{QA-based Evaluation Protocol.}
We adopt the SugarCrepe benchmark~\citep{hsieh2023sugarcrepe}, which is designed to test compositional understanding in vision-language models.
The benchmark provides an image paired with a correct and a perturbed caption, and is typically solved by selecting the caption with the higher image-text similarity score.
Since MLLMs are not assessed via simple similarity scores, we reformulate the benchmark as a question-answering task.
The model is prompted with an image and a multiple-choice query:

\begin{promptbox}[title=Judge prompt (example)]
\small
Which caption correctly describes the image?\\
A.~The horse is eating the grass.\\
B.~The grass is eating the horse.\\
Answer with a single letter (A or B).
\end{promptbox}

\paragraph{Models and Procedure.}
We evaluate recent open-source MLLMs: LLaVA-1.6~\citep{liu2024llavanext}, LLaMA-3.2-Vision~\citep{grattafiori2024llama}, the Qwen series~\citep{bai2025qwen3, qwen3.5}, and the InternVL series~\citep{zhu2025internvl3exploringadvancedtraining, wang2025internvl3}.
For comparison, we also report MLLM-based embedding models~\cite{jiang2024vlm2vec, meng2025vlm2vec,qwen3vlembedding} on the same benchmark.
To account for position bias, MLLMs are evaluated twice by swapping the position of the ground-truth caption.

Table~\ref{tab:pilot} shows that MLLMs achieve consistently high compositional ability, while their embedding-tuned counterparts lag substantially.
These results show that embedding-style fine-tuning weakens compositional understanding regardless of the underlying MLLM.
We therefore directly distill the compositional understanding of base MLLMs.

\section{Method}
\begin{figure*}
    \centering
    \includegraphics[width=\linewidth]{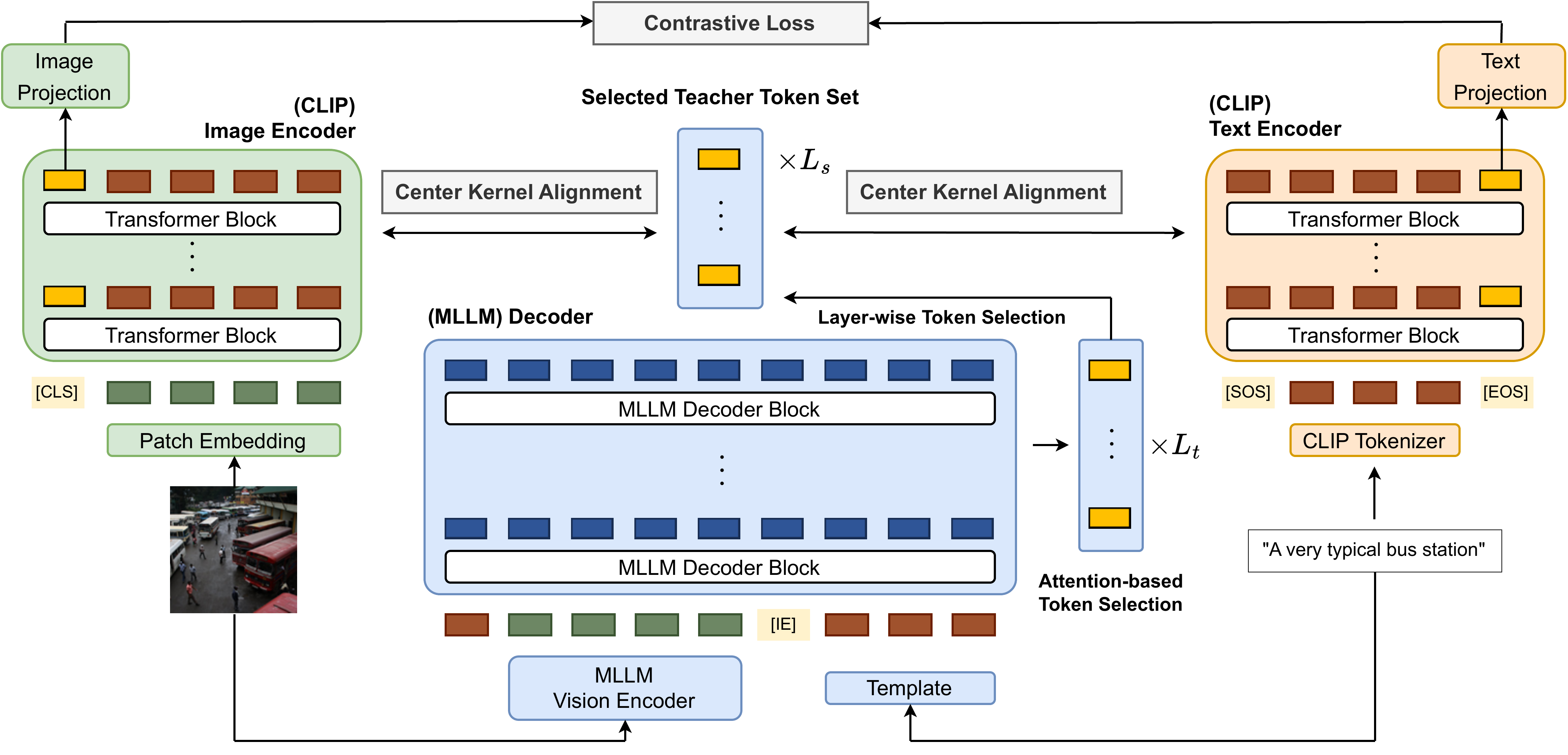}
    \caption{Overview of the MLLMCLIP framework. Student features are passed through auxiliary layers before alignment, and attention-based token selection obtains salient teacher tokens.}
    \label{fig:framework}
\end{figure*}

We now describe MLLMCLIP, a teacher-agnostic feature-level distillation framework that transfers multimodal knowledge from an MLLM teacher into a CLIP student, illustrated in Figure~\ref{fig:framework}.

\subsection{Architecture}
\subsubsection{Student Embedding (CLIP)}
The student model consists of separate CLIP-based image and text encoders, denoted as $f^I$ and $f^T$, respectively.
We follow the standard CLIP-style input processing: patch embedding followed by prepending a \texttt{[CLS]} token for image input $x^I$, and tokenization with \texttt{[SOS]} and \texttt{[EOS]} for text input~$x^T$.
Each sequence is passed through its encoder to produce layer-wise hidden states:
\begin{equation}
    h^I_l \in \mathbb{R}^{d_s^I}, \quad h^T_l \in \mathbb{R}^{d_s^T}, \quad l = 1, \dots, L_s,
\end{equation}
where $h^I_l$ and $h^T_l$ denote the hidden states of the \texttt{[CLS]} and \texttt{[EOS]} tokens from the $l$-th layer of the image and text encoders, respectively.
Here, $L_s$ is the number of encoder layers, and $d_s^I$, $d_s^T$ are the hidden dimensions of the image and text encoders.
To align the feature dimensions of the teacher and student, the hidden states are passed through auxiliary layers ($\text{Aux}^I$, $\text{Aux}^T$) as follows:
\begin{equation}
    h^{\text{Stud},I}_l = \text{Aux}^I(h^I_l), \quad
    h^{\text{Stud},T}_l = \text{Aux}^T(h^T_l)
\end{equation}
where $h^{\text{Stud},I}_l, h^{\text{Stud},T}_l \in \mathbb{R}^D$.
These student representations are used to match the corresponding teacher representations.
In parallel, the final-layer representations are projected into a common embedding space for contrastive learning.
Following the original CLIP architecture, these projection heads, $\text{Proj}^I$ and $\text{Proj}^T$, are implemented as a single linear layer:
\begin{equation}
    z^I = \text{Proj}^I(h^I_{L_s}), \quad
    z^T = \text{Proj}^T(h^T_{L_s}),
\end{equation}
where $z^I, z^T \in \mathbb{R}^d$.

\subsubsection{Teacher Embedding (MLLM)}
The teacher model jointly encodes image $x^I$ and text $x^T$ as a unified sequence. The image is first tokenized via patch embedding and processed through a vision encoder, which is part of the MLLM architecture.
The text input is inserted into a fixed template designed to activate the model's compositionality, as detailed in Appendix~\ref{app:implementation}.
We concatenate the image features with the \texttt{[Image~End]} token, followed by the text tokens, before passing them through the decoder blocks.

The teacher model produces hidden states from each decoder layer \( l \in \{1, \dots, L_t\} \), denoted as
\begin{equation}
    [h^{\text{MLLM}}_{l,1}, \dots, h^{\text{MLLM}}_{l,N}], \quad h^{\text{MLLM}}_{l,n} \in \mathbb{R}^{d_t}.
\end{equation}
Note that $N$ is the sequence length, and $L_t$ and $d_t$ denote the number of decoder layers and the feature dimension of the teacher model, respectively.

\paragraph{Token-wise Selection.}
To identify a representative teacher token within the sequence, we use self-attention weights from each decoder layer.
Let $\mathbf{A}_l^{(k)} \in \mathbb{R}^{N \times N}$ denote the attention matrix from head $k \in \{1, \dots, K\}$ at layer $l$, where $K$ is the number of attention heads.
We begin by averaging the attention weights across all heads:
\begin{equation}
    \bar{\mathbf{A}}_l = \frac{1}{K} \sum_{k=1}^K \mathbf{A}_l^{(k)} \in \mathbb{R}^{N \times N}.
\end{equation}
We then compute the maximum attention each token receives across all query positions:
\begin{equation}
    a_{l,n} = \max_{i \in \{1, \dots, N\}} \bar{\mathbf{A}}_l[i, n],
\end{equation}
where $\bar{\mathbf{A}}_l[i, n]$ denotes the attention weight from token $i$ (as query) to token $n$ (as key). Finally, we select the token with the highest received attention:
\begin{equation}
    n_l^\ast = \operatorname*{argmax}_{n \in \{1, \dots, N\}} a_{l,n}, \quad
    h^{\text{Teach}}_l = h^{\text{MLLM}}_{l, n_l^\ast}.
\end{equation}
This strategy selects the most attended token in each layer, under the assumption that such tokens are likely to carry salient multimodal information.

\paragraph{Layer-wise Selection.}
While token selection is guided by attention, we empirically select a subset of teacher layers for distillation to reduce computational cost.
Let $\mathcal{S} = \{s_1, \dots, s_{L_s}\} \subseteq \{1, \dots, L_t\}$ denote the selected set of teacher layers, where $|\mathcal{S}| = L_s$.
We explore strategies such as selecting layers from fixed relative positions (e.g., sampling from early, middle, and late stages of the teacher) or using uniform stride intervals across the full depth.
This design is motivated by the characteristic of Transformer representations, where lower layers tend to capture fine-grained information and higher layers encode abstract multimodal semantics.
Based on empirical results, we adopt the stride-based selection strategy for all experiments.

\subsection{Loss Functions}
\subsubsection{Contrastive Loss}
Following the standard CLIP training~\citep{radford2021learning}, we use an InfoNCE loss to align image and text embeddings.
Let \( z_i^I \) and \( z_i^T \) denote the image and text embeddings for the $i$-th sample in a batch of size $B$.
The contrastive loss using image embeddings as anchors is given by:
\begin{equation}
\resizebox{0.85\linewidth}{!}{
$\displaystyle\mathcal{L}_{\text{contrast}}^I = \frac{1}{B} \sum_{i=1}^{B}
    -\log \frac{\exp(\text{sim}(z_i^I, z_i^T)/\tau)}
    {\sum_{j=1}^{B} \exp(\text{sim}(z_i^I, z_j^T)/\tau)},$}
\end{equation}
where \( \text{sim}(\cdot, \cdot) \) denotes cosine similarity and \( \tau \) is a temperature.
Similarly, we compute \( \mathcal{L}_{\text{contrast}}^T \) by treating text embeddings as anchors.
The final contrastive loss is given by:
\begin{equation}
\mathcal{L}_{\text{contrast}} = \frac{1}{2} \left( \mathcal{L}_{\text{contrast}}^I + \mathcal{L}_{\text{contrast}}^T \right).
\end{equation}

\subsubsection{Distillation Loss}
To align intermediate representations between the teacher and student models, we consider a set of similarity-based objectives for distillation.
These objectives fall into two main categories:

\begin{itemize}
    \item \textbf{Direct similarity loss:} such as mean squared error (MSE), which operates on a per-sample basis and compares feature vectors directly.
    \item \textbf{Relational similarity loss:} such as Centered Kernel Alignment (CKA)~\citep{kornblith2019similarity}, which assesses structural similarity by aligning either the sample-level Gram matrices or the feature-level covariance matrices.
\end{itemize}
\paragraph{Direct Similarity Loss.}
Direct similarity losses enforce a sample-wise correspondence between the teacher's and the student's intermediate representations.
Within a mini-batch size $B$, let $h_{l,i}^{\text{Stud},I}$ denote the student's image representation for the $i$-th sample from layer $l$, and $h_{s_l,i}^{\text{Teach},I}$ be the corresponding teacher's image representation from a selected teacher layer $s_l$.
For each layer, we compute the similarity loss between the student and teacher representations for both image and text, and sum across layers.
The resulting objective is averaged over the batch:
\begin{align}
\mathcal{L}_{\text{distill}}
= \frac{1}{B} &\sum_{i=1}^B \frac{1}{L_s} \sum_{l=1}^{L_s}
\frac{1}{2} \Big(
    \ell_{\text{direct}}(h_{l,i}^{\text{Stud},I}, h_{l,i}^{\text{Teach},I}) \nonumber\\
 & + \ell_{\text{direct}}(h_{l,i}^{\text{Stud},T}, h_{l,i}^{\text{Teach},T})
\Big),
\end{align}
where $L_s$ is the number of student layers.
In our experiments, we consider MSE, cosine similarity, KL divergence, and JS divergence as direct loss functions, denoted as $\ell_{\text{direct}}$.

\paragraph{Relational Similarity Loss.}
We primarily adopt the CKA-based loss, which is effective for transferring knowledge between different architectures~\citep{dasgupta2025improving}.
Unlike direct similarity losses, CKA captures the global structural alignment of representations across a batch of samples.
We first gather the layer-wise representations for all $B$ samples in a mini-batch into matrices.
For a given modality, let $\mathbf{X}_l \in \mathbb{R}^{B \times D}$ be the matrix of student hidden states from layer $l$, and $\mathbf{Y}_{s_l}\in \mathbb{R}^{B \times D}$ be the matrix of corresponding teacher hidden states from layer $s_l$.
CKA operates by comparing the Gram matrices of these centered feature matrices. First, the matrices are centered:
\begin{equation}
    \tilde{\mathbf{X}}_l = \mathbf{X}_l - \frac{1}{B}\mathbf{1}\mathbf{1}^\top\mathbf{X}_l, \quad
    \tilde{\mathbf{Y}}_{s_l} = \mathbf{Y}_{s_l} - \frac{1}{B}\mathbf{1}\mathbf{1}^\top\mathbf{Y}_{s_l},
\end{equation}
where $\mathbf{1}$ is a column vector of ones. The Gram matrices, $K_l\in\mathbb{R}^{B \times B}$ and $L_{s_l}\in\mathbb{R}^{B \times B}$, are then computed:
\begin{equation}
    \mathbf{K}_l = \tilde{\mathbf{X}}_l \tilde{\mathbf{X}}_l^\top, \quad
    \mathbf{L}_{s_l} = \tilde{\mathbf{Y}}_{s_l} \tilde{\mathbf{Y}}_{s_l}^\top.
\end{equation}
Finally, CKA is calculated as the normalized Frobenius inner product of these Gram matrices:
\begin{equation}
    \text{CKA}(\mathbf{X}_l, \mathbf{Y}_{s_l}) = \frac{\langle \mathbf{K}_l, \mathbf{L}_{s_l} \rangle_F}{\|\mathbf{K}_l\|_F \|\mathbf{L}_{s_l}\|_F},
\end{equation}
where $\|\cdot\|_F$ and $\langle \cdot, \cdot \rangle_F$ denote the Frobenius norm and Frobenius inner product, respectively.
The CKA similarity score is converted into a loss for a single layer, $\ell_{\text{CKA}}$.
We adopt a common variant that uses a square root, which can provide better gradient properties and create a more sensitive loss when the similarity is high.
The total CKA loss is calculated by averaging the single-layer losses across both image and text modalities and summing them over a predefined set of layers:
\begin{equation}
    \ell_{\text{CKA}}(\mathbf{X}_l, \mathbf{Y}_{s_l}) = 1 - \sqrt{\text{CKA}(\mathbf{X}_l, \mathbf{Y}_{s_l})},
\end{equation}
\begin{equation}
\mathcal{L}_{\text{distill}} = \frac{1}{L_s} \sum_{l = 1}^{L_s} \frac{1}{2} \left( \ell_{\text{CKA}}(\mathbf{X}_l^I, \mathbf{Y}_{s_l}^I) + \ell_{\text{CKA}}(\mathbf{X}_l^T, \mathbf{Y}_{s_l}^T) \right).
\end{equation}
Using a weighting factor $\lambda$, the total training objective is defined as:
\begin{equation}
\mathcal{L}_{\text{total}} = \mathcal{L}_{\text{contrast}} + \lambda \mathcal{L}_{\text{distill}}.
\end{equation}

\begin{table*}[t]
\centering
\caption{Performance on 11 compositionality benchmarks.}
\vspace{-0.5em}
\renewcommand{\arraystretch}{1.1}
\resizebox{\textwidth}{!}{%
\begin{tabular}{ll|*{11}{r}|r}
\toprule
Method            & External Source
  & \rotatebox{90}{ARO}
  & \rotatebox{90}{CREPE}
  & \rotatebox{90}{EQBEN}
  & \rotatebox{90}{ImageCODE}
  & \rotatebox{90}{SugarCrepe}
  & \rotatebox{90}{SvoProbes}
  & \rotatebox{90}{VALSE}
  & \rotatebox{90}{VLChecklist}
  & \rotatebox{90}{WhatsUp}
  & \rotatebox{90}{Winoground}
  & \rotatebox{90}{SPEC}
  & \rotatebox{90}{Average} \\
\midrule
CLIP               & –            & 35.6 & 11.6 & 14.8 & 15.0 & 59.7 & 75.8 & 55.1 & 64.5 & 41.3 & 7.00 & 28.5 & 37.1 \\
LaCLIP             & LLAMA-7B        & 35.2 &  9.98 & 14.4 & 14.8 & 64.0 & 76.9 & 56.6 & 64.8 & 41.5 & 4.75 & 28.3 & 37.5 \\
NegCLIP            & Qwen3-4B        & 36.2 & 11.7  & 14.8 & 15.8 & 62.4 & 76.3 & 56.1 & 64.0 & 40.9 & 7.00 & 28.2 & 37.6 \\
FSC-CLIP           & Qwen3-4B        & 35.1 &  9.86 & 16.3 & 14.9 & 63.8 & 78.8 & 57.4 & 64.4 & 41.8 & 5.50 & 29.2 & 38.0 \\
TripletCLIP        & Qwen3-4B, SDXL-Turbo    & 34.6 & 10.5  & 16.4 & 16.9 & 65.7 & 78.8 & \textbf{59.4} & 64.8 & 41.1 & 4.25 & 29.0 & 38.3 \\
\cdashline{1-14}
\multirow{5}{*}{\shortstack{MLLMCLIP\\(Ours)}}
                   & LLaVA-1.6-Mistral-7B   & 32.1 & 10.7 & 15.5 & 15.6 & 69.3 & 79.4 & 55.9 & 64.6 & 43.6 & 7.25 & 26.5 & 38.2 \\
                   & Qwen3-VL-2B            & 34.6 & 12.0 & 17.0 & 17.7 & 71.3 & \textbf{82.3} & 59.2 & \textbf{66.9} & \textbf{43.8} & \textbf{8.00} & 28.2 & 40.1 \\
                   & Qwen3.5-2B             & 34.1 & 11.9 & 17.9 & 17.3 & \textbf{72.1} & 81.4 & 58.8 & 66.3 & 43.7 & 7.75 & 28.3 & 40.0 \\
                   & InternVL3-2B           & 35.0 & \textbf{12.3} & 18.1 & 17.6 & 72.0 & 81.8 & 57.9 & 66.2 & 43.5 & 6.00 & \textbf{29.8} & 40.0 \\
                   & InternVL3.5-2B         & \textbf{37.0} & 12.2 & \textbf{18.7} & \textbf{18.3} & 71.8 & 81.9 & 58.3 & \textbf{66.9} & 43.5 & 7.00 & 29.1 & \textbf{40.4} \\
\bottomrule
\end{tabular}%
}
\label{tab:main_comp}
\end{table*}

\section{Experiments}
\paragraph{Data \& Pre-processing.}
We use CC3M~\citep{sharma2018conceptual} as the base pretraining dataset.
For each image-caption pair, we format the caption with a structured template designed to encourage compositional reasoning and pass it through the frozen MLLM teacher to extract per-layer features offline.
For evaluation, we use 11 compositionality benchmarks, 13 zero-shot classification datasets, and 2 image-text retrieval benchmarks.
Full benchmark descriptions and the teacher prompt template are provided in Appendix~\ref{app:dataset} and~\ref{app:implementation}.

\paragraph{Implementation Details.}
We evaluate MLLMCLIP with five MLLM teachers: LLaVA-1.6-Mistral-7B~\citep{liu2024llavanext}, Qwen3-VL-2B~\citep{bai2025qwen3}, Qwen3.5-2B~\citep{qwen3.5}, InternVL3-2B~\citep{zhu2025internvl3exploringadvancedtraining}, and InternVL3.5-2B~\citep{wang2025internvl3}.
Each teacher is distilled into a CLIP Base student, comprising a ViT-B/32 image encoder and a 12-layer Transformer text encoder, with an auxiliary head of a linear projection followed by Layer Normalization on each modality branch.
Every student is trained from scratch on CC3M rather than fine-tuned from a pretrained CLIP checkpoint.
All models are trained with a global batch size of 4096, and any additional negative samples introduced by competing methods are included within the same budget.
Full hyperparameter configurations are provided in Appendix~\ref{app:implementation}.

\subsection{Reproducing Prior Works}
We reproduce four CLIP-enhancement baselines: LaCLIP~\citep{fan2023improving}, NegCLIP~\citep{yuksekgonul2022and}, FSC-CLIP~\citep{oh2024preserving}, and TripletCLIP~\citep{patel2024tripletclip}.
These methods enhance CLIP training by incorporating supervision from external Large Language Models and image generative models.
For LaCLIP, we use the publicly released LLaMA-generated positive captions\footnote{\url{https://github.com/LijieFan/LaCLIP}}.
For NegCLIP and FSC-CLIP, we generate hard negative captions using Qwen3-4B~\citep{yang2025qwen3}.
For TripletCLIP, we feed the Qwen3-4B-generated captions into SDXL-Turbo~\citep{sauer2024adversarial} to synthesize corresponding negative images using the prompt templates from the original paper, yielding one negative caption-image pair per CC3M sample.

\subsection{Main Results}
\paragraph{Compositionality.}
\begin{table*}[t]
  \centering
  \caption{Zero-shot classification performance on 13 classification datasets.}
  \vspace{-0.5em}
  \renewcommand{\arraystretch}{1.1}
  \resizebox{\textwidth}{!}{%
    \begin{tabular}{ll|*{13}{r}|r}
      \toprule
      Method & External Source
       & \rotatebox{90}{Caltech\,101}
       & \rotatebox{90}{CIFAR-10}
       & \rotatebox{90}{CIFAR-100}
       & \rotatebox{90}{DTD}
       & \rotatebox{90}{EuroSAT}
       & \rotatebox{90}{FER\,2013}
       & \rotatebox{90}{Flower\,102}
       & \rotatebox{90}{Food\,101}
       & \rotatebox{90}{ImageNet}
       & \rotatebox{90}{KITTI}
       & \rotatebox{90}{Pet}
       & \rotatebox{90}{RESISC45}
       & \rotatebox{90}{VOC\,2007}
       & \rotatebox{90}{Average} \\
      \midrule
      CLIP          & –           & 37.8 & 53.3 & 21.7 &  7.93 & 12.1 & 12.4 &  8.88 &  8.91 & 12.4 & 20.0 &  9.95 & 15.9 & 54.0 & 21.2 \\
      LaCLIP        & LLAMA-7B       & 41.7 & 49.7 & 21.2 & 11.2  & 14.3 & 19.5 & 11.2  &  9.34 & 13.7 & 30.7 & 10.1  & 17.1 & 60.9 & 23.9 \\
      NegCLIP       & Qwen3-4B       & 41.4 & 55.3 & 24.1 & 10.6  & 18.3 & 18.9 &  9.16 &  9.49 & 13.7 & 30.4 &  8.81 & 17.9 & 56.8 & 24.2 \\
      FSC-CLIP       & Qwen3-4B       & 41.5 & 54.2 & 25.1 &  9.84 & 18.0 & 16.8 &  9.03 &  9.40 & 13.5 & 20.8 & 10.1 & 20.3 & 63.1 & 24.0 \\
      TripletCLIP   & Qwen3-4B, SDXL-Turbo   & 42.4 & 46.5 & 22.3 & 13.5  & 24.6 & 17.2 & 10.4  &  9.97 & 14.2 & 23.9 & 10.8  & 22.8 & 56.3 & 24.2 \\
      \cdashline{1-16}
      \multirow{5}{*}{\shortstack{MLLMCLIP\\(Ours)}}
                    & LLaVA-1.6-Mistral-7B   & 48.2 & 65.0 & 32.1 & 13.8 & 19.0 & 14.8 & 8.4 & 10.8 & 15.4 & 33.2 & 10.2 & 24.4 & 66.6 & 27.8 \\
                    & Qwen3-VL-2B            & 48.8 & 67.5 & 33.0 & 15.0 & 25.5 & 19.1 & 10.3 & 12.3 & 16.7 & 43.2 & 12.3 & 19.6 & 64.1 & 29.8 \\
                    & Qwen3.5-2B             & \textbf{51.3} & 67.6 & 34.6 & 16.2 & \textbf{26.1} & \textbf{23.5} & 9.13 & 12.4 & 17.5 & 39.1 & 13.8 & 24.3 & \textbf{71.8} & 31.3 \\
                    & InternVL3-2B           & 48.9 & \textbf{71.4} & 34.8 & \textbf{16.9} & 22.2 & 19.7 & 10.7 & \textbf{12.8} & 17.5 & 46.7 & \textbf{16.5} & 23.3 & 66.8 & \textbf{31.4} \\
                    & InternVL3.5-2B         & 51.0 & 69.4 & \textbf{36.6} & 16.8 & 17.9 & 19.8 & \textbf{11.4} & 12.0 & \textbf{17.9} & \textbf{47.6} & 13.6 & \textbf{25.5} & 68.1 & 31.3 \\
      \bottomrule
    \end{tabular}%
  }
  \label{tab:main_zs}
\end{table*}

\begin{table*}[t]
  \centering
  \caption{Zero-shot retrieval performance on MSCOCO and Flickr-30K datasets.}
  \vspace{-0.5em}
  \renewcommand{\arraystretch}{1.1}
  \resizebox{\textwidth}{!}{%
    \begin{tabular}{ll|*{12}{r}}
      \toprule
        & & \multicolumn{6}{c}{Image–to–text retrieval}  & \multicolumn{6}{c}{Text–to–image retrieval} \\
        \cmidrule(lr){3-8}\cmidrule(lr){9-14}
        & & \multicolumn{3}{c}{MSCOCO} & \multicolumn{3}{c}{Flickr-30K} & \multicolumn{3}{c}{MSCOCO} & \multicolumn{3}{c}{Flickr-30K} \\
        \cmidrule(lr){3-5}\cmidrule(lr){6-8}\cmidrule(lr){9-11}\cmidrule(lr){12-14}
        Method & External Source
        & R@1 & R@5 & R@10
        & R@1 & R@5 & R@10
        & R@1 & R@5 & R@10
        & R@1 & R@5 & R@10 \\
          \midrule
          CLIP               & –          &  9.02 & 24.3 & 33.7 & 18.3 & 39.9 & 50.4 &  6.85 & 18.7 & 27.1 & 12.9 & 31.6 & 41.5 \\
          LaCLIP             & LLAMA-7B      &  9.80 & 24.8 & 33.7 & 20.8 & 40.4 & 51.5 &  6.72 & 19.1 & 27.8 & 15.9 & 36.8 & 47.8 \\
          NegCLIP       & Qwen3-4B      &  9.38 & 24.4 & 34.4 & 20.1 & 42.8 & 53.6 &  7.66 & 20.0 & 28.5 & 13.7 & 32.2 & 42.7 \\
          FSC-CLIP      & Qwen3-4B      & 11.4  & 27.8 & 37.7 & 22.0 & 46.7 & 58.0 &  8.91 & 23.5 & 32.8 & 17.4 & 38.4 & 49.0 \\
          TripletCLIP        & Qwen3-4B, SDXL-Turbo  & 12.3  & 29.9 & 40.3 & 22.9 & 46.8 & 59.6 &  9.11 & 23.5 & 33.1 & 17.8 & 39.1 & 49.8 \\
          \cdashline{1-14}
          \multirow{5}{*}{\shortstack{MLLMCLIP\\(Ours)}}
                   & LLaVA-1.6-Mistral-7B   & 11.2 & 31.9 & 43.7 & 26.2 & 56.8 & 66.6 & 9.4 & 26.4 & 36.5 & 21.3 & 45.0 & 55.4 \\
                   & Qwen3-VL-2B            & 12.6 & 32.7 & 43.4 & 28.9 & 56.1 & 67.0 & 11.2 & 27.8 & 37.7 & 22.3 & 46.1 & 57.1 \\
                   & Qwen3.5-2B             & 13.8 & 34.3 & 45.7 & 28.6 & 58.2 & 67.5 & 11.4 & 28.4 & 38.9 & 22.7 & 46.8 & 57.0 \\
                   & InternVL3-2B           & \textbf{14.7} & \textbf{35.2} & 47.0 & \textbf{31.2} & \textbf{59.4} & \textbf{69.0} & 11.6 & 29.1 & 39.0 & \textbf{23.8} & \textbf{47.5} & \textbf{58.5} \\
                   & InternVL3.5-2B         & 14.4 & 35.0 & \textbf{47.1} & 29.8 & 57.7 & 67.9 & \textbf{11.7} & \textbf{29.2} & \textbf{39.2} & 23.1 & 47.2 & 57.9 \\
          \bottomrule
    \end{tabular}%
  }
  \label{tab:zs_retrieval_full}
\end{table*}

Table~\ref{tab:main_comp} reports performance on the 11 compositionality benchmarks.
Previous CLIP-enhancement methods show modest gains over CLIP.
LaCLIP, which adds positive captions, yields a marginal improvement.
NegCLIP and FSC-CLIP achieve a slightly larger gain with LLM-generated hard negative captions.
Even with image synthesis, TripletCLIP brings limited further benefit over caption-only methods.
In contrast, MLLMCLIP outperforms previous methods, with the gain becoming pronounced for stronger teachers.
This confirms that compositional reasoning is more effectively strengthened by feature-level distillation from MLLM than by data-level pipelines that synthesize additional training samples.
To verify that this gain stems from the transfer mechanism rather than from teacher identity, Appendix~\ref{app:zs} repeats the comparison with the same MLLM supplying both the data-level and the feature-level supervision, and separately ablates the teacher prompt.

\paragraph{Zero-shot Classification \& Retrieval.}
Tables~\ref{tab:main_zs} and~\ref{tab:zs_retrieval_full} report zero-shot classification on 13 datasets and image-text retrieval on 2 datasets.
Previous methods bring moderate improvements over CLIP, suggesting that their supervision primarily targets compositionality rather than general-purpose recognition or retrieval.
Across all five teachers, MLLMCLIP improves over every previous method on both axes.
This confirms that distilling from MLLMs strengthens compositional and general vision-language ability simultaneously.

\subsection{Ablation Studies}
All ablations use Qwen3.5-2B as the teacher, the most recent MLLM in our comparison.
We report results across three evaluation groups. Compositionality is measured as the average accuracy over 11 datasets and denoted as \textbf{Comp.}. For zero-shot classification, we report the average performance over 13 datasets as \textbf{Zero-shot Cls.}. For retrieval, we report the average Recall@1 over image-to-text and text-to-image retrieval on the two datasets, denoted as \textbf{Ret.}
Ablations on the teacher layer selection, the distillation weight, and the teacher prompt, together with a comparison that controls for teacher identity, are deferred to Appendix~\ref{app:zs}.

\paragraph{Teacher Token Selection.}
Table~\ref{tab:4_abl_token} compares strategies for choosing which teacher tokens to distill.
We first evaluate fixed positions, the end of the image segment (\texttt{[Image End]}) and the end of the full multimodal sequence (\texttt{[Text End]}), which under causal masking carry unimodal visual and multimodal understanding, respectively.
Both improve over the baseline, with gains accumulating as the signal becomes multimodal and as the text branch is also supervised.
In contrast, attention-based selection outperforms every fixed position on all metrics, showing the benefit of adaptively choosing informative tokens from each layer.
Varying how many attention-selected tokens are kept per layer, from attention-weighted pooling and top-$k$ pooling down to a single token, changes little, indicating that what matters is locating the teacher signal by attention rather than how many tokens are aggregated.
We therefore keep a single token as the default for its simplicity and its one-vector-per-layer feature store.
\begin{table}[t]
    \centering
    \caption{Effect of teacher token selection and per-layer aggregation. \texttt{[Image End]} and \texttt{[Text End]} denote the token positions at the end of the image segment and of the full multimodal sequence. The lower block varies how many attention-selected tokens are kept per layer.}
    \small
    \resizebox{\linewidth}{!}{
        \begin{tabular}{ll|ccc}
            \toprule
            \textbf{Image Teacher} & \textbf{Text Teacher} & \textbf{Comp.} & \textbf{Zero-shot Cls.} & \textbf{Ret.} \\
            \midrule
            None                     & None            & 37.1 & 21.2 & 11.8 \\
            \texttt{[Image End]}       & None            & 37.5 & 25.2 & 16.2 \\
            \texttt{[Text End]}        & None            & 38.0 & 26.6 & 17.8 \\
            \texttt{[Text End]}        & \texttt{[Text End]}        & 38.2 & 28.1 & 17.8 \\
            \midrule
            \multicolumn{2}{l|}{Attention-weighted pooling}          & \textbf{40.2} & 31.1 & 18.9 \\
            \multicolumn{2}{l|}{Top-$k$ multi-token pooling ($k=8$)} & 39.7 & 30.8 & 18.2 \\
            \multicolumn{2}{l|}{Single-token selection (default)}    & 40.0 & \textbf{31.3} & \textbf{19.1} \\
            \bottomrule
        \end{tabular}}
    \label{tab:4_abl_token}
\end{table}

\paragraph{Loss Functions.}
\begin{table}[t]
    \centering
    \caption{Effect of different distillation loss functions on downstream performance. }
    \small
    \resizebox{\linewidth}{!}{
        \begin{tabular}{ll|ccc}
            \toprule
            \textbf{Type} & \textbf{Loss} & \textbf{Comp.} & \textbf{Zero-shot Cls.} & \textbf{Ret.} \\
            \midrule
            \multirow{4}{*}{Direct}&MSE              & 30.2 & 7.00 & 0.05 \\
            &Cosine           & 38.6 & 27.5 & 17.2 \\
            &KL Div.    & 38.2 & 28.1 & 18.4 \\
            &JS Div.    & 37.4 & 25.8 & 16.0 \\
            \midrule
            Relational &CKA       & \textbf{40.0} & \textbf{31.3} & \textbf{19.1} \\
            \bottomrule
        \end{tabular}
    }
    \label{tab:4_abl_loss}
\end{table}

Table~\ref{tab:4_abl_loss} compares distillation losses.
MSE loss enforces a strict element-wise match between feature vectors, resulting in performance collapse.
This stems from the fundamental architectural mismatch between the generative decoder and the representational encoder.
Scale-normalized objectives alleviate this issue and stabilize training, but remain suboptimal since they focus only on pointwise alignment.
Notably, CKA consistently outperforms direct similarity approaches across all downstream tasks.
This suggests that preserving structural relationships among samples is more effective for knowledge transfer than per-sample feature matching, especially in a cross-architecture setting.

\paragraph{Preprocessing Efficiency.}
\begin{table}[t]
    \centering
    \caption{Per-batch preprocessing time at batch size 32, comparing data synthesis against feature extraction. Mean $\pm$ std over 5 batches on a single GPU.}
    \vspace{-0.25em}
    \small
    \resizebox{\linewidth}{!}{
        \begin{tabular}{l|cc}
            \toprule
                              & Data & Feature \\
            \midrule
            Text gen.\ (s)    & $0.278 \pm 0.001$ & -- \\
            Image gen.\ (s)   & $0.557 \pm 0.012$ & -- \\
            Feature ext.\ (s) & --                & $0.232 \pm 0.005$ \\
            \midrule
            Total (s)         & $0.835 \pm 0.012$ & $\mathbf{0.232 \pm 0.005}$ \\
            \bottomrule
        \end{tabular}}
    \label{tab:eff_pre}
\end{table}

Table~\ref{tab:eff_pre} compares the preprocessing cost of feature-level and data-level distillation.
Generating negative captions and images using a cascaded pipeline takes roughly $3.6\times$ the per-batch time required for feature extraction.
Pre-extracted teacher features also enable larger student batch sizes under the same GPU memory budget, since the student avoids encoding the additional negatives that data-level methods feed in during training.

\subsection{Comparison with MLLM-based Embedding Models}\label{sec:vs_embedding}
\begin{table}[t]
    \centering
    \caption{Single-sample inference cost and SugarCrepe accuracy. Mean $\pm$ std over 10 runs on a single GPU.}
    \vspace{-0.25em}
    \small
    \resizebox{\linewidth}{!}{
        \begin{tabular}{l|cc|c}
            \toprule
            Model & \#Params (B) & Inf.\ time (ms) & SugarCrepe \\
            \midrule
            VLM2Vec-v1  & 2.21 & $114.60 \pm 0.33$ & 64.6 \\
            VLM2Vec-v2  & 2.22 & $115.92 \pm 0.23$ & 72.7 \\
            Qwen3-VL-Emb.-2B & 2.13 & $52.36 \pm 0.21$  & \textbf{83.0} \\
            MLLMCLIP              & \textbf{0.15} & $\mathbf{4.57 \pm 0.04}$ & 71.3 \\
            \bottomrule
        \end{tabular}}
    \label{tab:eff_inf}
\end{table}

A recent line of work directly fine-tunes an MLLM into an embedding model, such as VLM2Vec~\citep{jiang2024vlm2vec} and Qwen3-VL-Embedding~\citep{bai2025qwen3}.
These models retain the full MLLM backbone at inference, requiring an order of magnitude more parameters and latency than a CLIP-style dual-encoder.
MLLMCLIP closes this gap from the opposite direction: it distills MLLM knowledge into a CLIP student rather than turning the MLLM itself into an embedding model.
As shown in Table~\ref{tab:eff_inf}, this yields a model that comes close to VLM2Vec-v2 on SugarCrepe at roughly $25\times$ lower latency.
Qwen3-VL-Embedding attains higher accuracy but at $11\times$ the inference cost of MLLMCLIP, leaving a clear trade-off between accuracy and efficiency.
The pilot study in Section~\ref{sec:pilot} shows that these embedding models still fall short of the base MLLMs from which they are tuned.

\section{Conclusion}
In this work, we present MLLMCLIP, a feature-level distillation framework that transfers multimodal knowledge from an MLLM teacher into a CLIP student.
Our framework addresses the fundamental architectural mismatch between the two through attention-based teacher-token selection and a structure-aware CKA distillation loss.
MLLMCLIP achieves strong performance across compositionality, zero-shot classification, and image-text retrieval, demonstrating the potential of MLLM-guided distillation for building lightweight vision-language encoders.
The framework is teacher-agnostic, applying to a wide range of MLLM teachers as shown by consistent gains across diverse teacher families.

\section*{Limitations}
While our work demonstrates the effectiveness of MLLM-to-CLIP distillation, it has a few limitations.
The capabilities of the teacher MLLM cap the performance of MLLMCLIP: biases or reasoning failures in the teacher can transfer to the student during distillation.
Although we propose a rigorous protocol for teacher selection, the optimal teacher may vary across downstream tasks.
Finally, we trade repeated runs for teacher diversity: each configuration is reported single-seed, but the gains remain consistent across all five MLLM teachers and across all benchmarks.

\bibliography{custom}

@article{hsieh2023sugarcrepe,
  title={Sugarcrepe: Fixing hackable benchmarks for vision-language compositionality},
  author={Hsieh, Cheng-Yu and Zhang, Jieyu and Ma, Zixian and Kembhavi, Aniruddha and Krishna, Ranjay},
  journal={Advances in neural information processing systems},
  volume={36},
  pages={31096--31116},
  year={2023}
}

@misc{liu2024llavanext,
    title={LLaVA-NeXT: Improved reasoning, OCR, and world knowledge},
    url={https://llava-vl.github.io/blog/2024-01-30-llava-next/},
    author={Liu, Haotian and Li, Chunyuan and Li, Yuheng and Li, Bo and Zhang, Yuanhan and Shen, Sheng and Lee, Yong Jae},
    month={January},
    year={2024}
}

@misc{zhu2025internvl3exploringadvancedtraining,
      title={InternVL3: Exploring Advanced Training and Test-Time Recipes for Open-Source Multimodal Models},
      author={Jinguo Zhu and Weiyun Wang and Zhe Chen and Zhaoyang Liu and Shenglong Ye and Lixin Gu and Hao Tian and Yuchen Duan and Weijie Su and Jie Shao and Zhangwei Gao and Erfei Cui and Xuehui Wang and Yue Cao and Yangzhou Liu and Xingguang Wei and Hongjie Zhang and Haomin Wang and Weiye Xu and Hao Li and Jiahao Wang and Nianchen Deng and Songze Li and Yinan He and Tan Jiang and Jiapeng Luo and Yi Wang and Conghui He and Botian Shi and Xingcheng Zhang and Wenqi Shao and Junjun He and Yingtong Xiong and Wenwen Qu and Peng Sun and Penglong Jiao and Han Lv and Lijun Wu and Kaipeng Zhang and Huipeng Deng and Jiaye Ge and Kai Chen and Limin Wang and Min Dou and Lewei Lu and Xizhou Zhu and Tong Lu and Dahua Lin and Yu Qiao and Jifeng Dai and Wenhai Wang},
      year={2025},
      eprint={2504.10479},
      archivePrefix={arXiv},
      primaryClass={cs.CV},
      url={https://arxiv.org/abs/2504.10479},
}

@article{grattafiori2024llama,
  title={The llama 3 herd of models},
  author={Grattafiori, Aaron and Dubey, Abhimanyu and Jauhri, Abhinav and Pandey, Abhinav and Kadian, Abhishek and Al-Dahle, Ahmad and Letman, Aiesha and Mathur, Akhil and Schelten, Alan and Vaughan, Alex and others},
  journal={arXiv preprint arXiv:2407.21783},
  year={2024}
}

@inproceedings{dasgupta2025improving,
  title={Improving language model distillation through hidden state matching},
  author={Dasgupta, Sayantan and Cohn, Trevor},
  booktitle={The Thirteenth International Conference on Learning Representations},
  year={2025}
}

@inproceedings{radford2021learning,
  title={Learning transferable visual models from natural language supervision},
  author={Radford, Alec and Kim, Jong Wook and Hallacy, Chris and Ramesh, Aditya and Goh, Gabriel and Agarwal, Sandhini and Sastry, Girish and Askell, Amanda and Mishkin, Pamela and Clark, Jack and others},
  booktitle={International conference on machine learning},
  pages={8748--8763},
  year={2021},
  organization={PmLR}
}

@inproceedings{lai2024veclip,
  title={Veclip: Improving clip training via visual-enriched captions},
  author={Lai, Zhengfeng and Zhang, Haotian and Zhang, Bowen and Wu, Wentao and Bai, Haoping and Timofeev, Aleksei and Du, Xianzhi and Gan, Zhe and Shan, Jiulong and Chuah, Chen-Nee and others},
  booktitle={European Conference on Computer Vision},
  pages={111--127},
  year={2024},
  organization={Springer}
}

@article{fan2023improving,
  title={Improving clip training with language rewrites},
  author={Fan, Lijie and Krishnan, Dilip and Isola, Phillip and Katabi, Dina and Tian, Yonglong},
  journal={Advances in Neural Information Processing Systems},
  volume={36},
  pages={35544--35575},
  year={2023}
}

@article{patel2024tripletclip,
  title={Tripletclip: Improving compositional reasoning of clip via synthetic vision-language negatives},
  author={Patel, Maitreya and Kusumba, Naga Sai Abhiram and Cheng, Sheng and Kim, Changhoon and Gokhale, Tejas and Baral, Chitta and others},
  journal={Advances in neural information processing systems},
  volume={37},
  pages={32731--32760},
  year={2024}
}

@article{oh2024preserving,
  title={Preserving Multi-Modal Capabilities of Pre-trained VLMs for Improving Vision-Linguistic Compositionality},
  author={Oh, Youngtaek and Cho, Jae Won and Kim, Dong-Jin and Kweon, In So and Kim, Junmo},
  journal={arXiv preprint arXiv:2410.05210},
  year={2024}
}

@article{yuksekgonul2022and,
  title={When and why vision-language models behave like bags-of-words, and what to do about it?},
  author={Yuksekgonul, Mert and Bianchi, Federico and Kalluri, Pratyusha and Jurafsky, Dan and Zou, James},
  journal={arXiv preprint arXiv:2210.01936},
  year={2022}
}

@incollection{fellbaum2010wordnet,
  title={WordNet},
  author={Fellbaum, Christiane},
  booktitle={Theory and applications of ontology: computer applications},
  pages={231--243},
  year={2010},
  publisher={Springer}
}

@inproceedings{sharma2018conceptual,
  title = {Conceptual Captions: A Cleaned, Hypernymed, Image Alt-text Dataset For Automatic Image Captioning},
  author = {Sharma, Piyush and Ding, Nan and Goodman, Sebastian and Soricut, Radu},
  booktitle = {Proceedings of ACL},
  year = {2018},
}

@inproceedings{ma2023crepe,
  title={Crepe: Can vision-language foundation models reason compositionally?},
  author={Ma, Zixian and Hong, Jerry and Gul, Mustafa Omer and Gandhi, Mona and Gao, Irena and Krishna, Ranjay},
  booktitle={Proceedings of the IEEE/CVF Conference on Computer Vision and Pattern Recognition},
  pages={10910--10921},
  year={2023}
}

@article{parcalabescu2021valse,
  title={VALSE: A task-independent benchmark for vision and language models centered on linguistic phenomena},
  author={Parcalabescu, Letitia and Cafagna, Michele and Muradjan, Lilitta and Frank, Anette and Calixto, Iacer and Gatt, Albert},
  journal={arXiv preprint arXiv:2112.07566},
  year={2021}
}

@article{zhao2022vl,
  title={Vl-checklist: Evaluating pre-trained vision-language models with objects, attributes and relations},
  author={Zhao, Tiancheng and Zhang, Tianqi and Zhu, Mingwei and Shen, Haozhan and Lee, Kyusong and Lu, Xiaopeng and Yin, Jianwei},
  journal={arXiv preprint arXiv:2207.00221},
  year={2022}
}

@article{kamath2023s,
  title={What's" up" with vision-language models? investigating their struggle with spatial reasoning},
  author={Kamath, Amita and Hessel, Jack and Chang, Kai-Wei},
  journal={arXiv preprint arXiv:2310.19785},
  year={2023}
}

@article{krojer2022image,
  title={Image retrieval from contextual descriptions},
  author={Krojer, Benno and Adlakha, Vaibhav and Vineet, Vibhav and Goyal, Yash and Ponti, Edoardo and Reddy, Siva},
  journal={arXiv preprint arXiv:2203.15867},
  year={2022}
}

@inproceedings{hendricks2021probing,
  title={Probing image-language transformers for verb understanding},
  author={Hendricks, Lisa Anne and Nematzadeh, Aida},
  booktitle={Findings of the Association for Computational Linguistics: ACL-IJCNLP 2021},
  pages={3635--3644},
  year={2021}
}

@inproceedings{thrush2022winoground,
  title={Winoground: Probing vision and language models for visio-linguistic compositionality},
  author={Thrush, Tristan and Jiang, Ryan and Bartolo, Max and Singh, Amanpreet and Williams, Adina and Kiela, Douwe and Ross, Candace},
  booktitle={Proceedings of the IEEE/CVF Conference on Computer Vision and Pattern Recognition},
  pages={5238--5248},
  year={2022}
}

@inproceedings{wang2023equivariant,
  title={Equivariant similarity for vision-language foundation models},
  author={Wang, Tan and Lin, Kevin and Li, Linjie and Lin, Chung-Ching and Yang, Zhengyuan and Zhang, Hanwang and Liu, Zicheng and Wang, Lijuan},
  booktitle={Proceedings of the IEEE/CVF International Conference on Computer Vision},
  pages={11998--12008},
  year={2023}
}

@inproceedings{peng2024synthesize,
  title={Synthesize diagnose and optimize: Towards fine-grained vision-language understanding},
  author={Peng, Wujian and Xie, Sicheng and You, Zuyao and Lan, Shiyi and Wu, Zuxuan},
  booktitle={Proceedings of the IEEE/CVF Conference on Computer Vision and Pattern Recognition},
  pages={13279--13288},
  year={2024}
}

@inproceedings{mu2022slip,
  title={Slip: Self-supervision meets language-image pre-training},
  author={Mu, Norman and Kirillov, Alexander and Wagner, David and Xie, Saining},
  booktitle={European conference on computer vision},
  pages={529--544},
  year={2022},
  organization={Springer}
}

@article{li2021supervision,
  title={Supervision exists everywhere: A data efficient contrastive language-image pre-training paradigm},
  author={Li, Yangguang and Liang, Feng and Zhao, Lichen and Cui, Yufeng and Ouyang, Wanli and Shao, Jing and Yu, Fengwei and Yan, Junjie},
  journal={arXiv preprint arXiv:2110.05208},
  year={2021}
}

@article{lee2022uniclip,
  title={Uniclip: Unified framework for contrastive language-image pre-training},
  author={Lee, Janghyeon and Kim, Jongsuk and Shon, Hyounguk and Kim, Bumsoo and Kim, Seung Hwan and Lee, Honglak and Kim, Junmo},
  journal={Advances in Neural Information Processing Systems},
  volume={35},
  pages={1008--1019},
  year={2022}
}

@article{goel2022cyclip,
  title={Cyclip: Cyclic contrastive language-image pretraining},
  author={Goel, Shashank and Bansal, Hritik and Bhatia, Sumit and Rossi, Ryan and Vinay, Vishwa and Grover, Aditya},
  journal={Advances in Neural Information Processing Systems},
  volume={35},
  pages={6704--6719},
  year={2022}
}

@article{chen2015microsoft,
  title={Microsoft coco captions: Data collection and evaluation server},
  author={Chen, Xinlei and Fang, Hao and Lin, Tsung-Yi and Vedantam, Ramakrishna and Gupta, Saurabh and Doll{\'a}r, Piotr and Zitnick, C Lawrence},
  journal={arXiv preprint arXiv:1504.00325},
  year={2015}
}

@inproceedings{plummer2015flickr30k,
  title={Flickr30k entities: Collecting region-to-phrase correspondences for richer image-to-sentence models},
  author={Plummer, Bryan A and Wang, Liwei and Cervantes, Chris M and Caicedo, Juan C and Hockenmaier, Julia and Lazebnik, Svetlana},
  booktitle={Proceedings of the IEEE international conference on computer vision},
  pages={2641--2649},
  year={2015}
}

@article{fei2007learning, title={Learning generative visual models from few training examples: An incremental Bayesian approach tested on 101 object categories}, author={Fei-Fei, Li and Fergus, Rob and Perona, Pietro}, journal={Computer vision and Image understanding}, volume={106}, number={1}, pages={59--70}, year={2007}, publisher={Elsevier}}

@techreport{krizhevsky2009learning, title={Learning multiple layers of features from tiny images}, author={Krizhevsky, Alex}, year={2009}, institution={University of Toronto}}

@InProceedings{cimpoi14describing, Author = {M. Cimpoi and S. Maji and I. Kokkinos and S. Mohamed and A. Vedaldi}, Title = {Describing Textures in the Wild}, Booktitle = {Proceedings of the IEEE Conf. on Computer Vision and Pattern Recognition (CVPR)}, Year = {2014} }

@article{helber2019eurosat, title={EuroSAT: A Novel Dataset and Deep Learning Benchmark for Land Use and Land Cover Classification}, author={Helber, Patrick and Bischke, Benjamin and Dengel, Andreas and Borth, Damian}, journal={IEEE Journal of Selected Topics in Applied Earth Observations and Remote Sensing}, volume={12}, number={7}, pages={2217--2226}, year={2019}}

@inproceedings{goodfellow2013challenges, title={Challenges in representation learning: A report on three machine learning contests}, author={Goodfellow, Ian J and Erhan, Dumitru and Carrier, Pierre Luc and Courville, Aaron and Mirza, Mehdi and Hamner, Brandon and Cukierski, William and Tang, Yuan and Thaler, David and Lee, Dong-Hyun and others}, booktitle={International Conference on Neural Information Processing}, pages={117--124}, year={2013}, organization={Springer}}

@InProceedings{Nilsback08, author = "Nilsback, M-E. and Zisserman, A.", title = "Automated Flower Classification over a Large Number of Classes", booktitle = "Proceedings of the Indian Conference on Computer Vision, Graphics and Image Processing", year = "2008", month = "Dec" }

@inproceedings{bossard14, title={Food-101 -- Mining Discriminative Components with Random Forests}, author={Bossard, Lukas and Guillaumin, Matthieu and Van Gool, Luc}, booktitle={European Conference on Computer Vision}, year={2014}}

@inproceedings{deng2009imagenet, title={ImageNet: A Large-Scale Hierarchical Image Database}, author={Deng, Jia and Dong, Wei and Socher, Richard and Li, Li-Jia and Li, Kai and Fei-Fei, Li}, booktitle={IEEE conference on computer vision and pattern recognition}, pages={248--255}, year={2009}, organization={Ieee}}

@article{geiger2013vision, title={Vision meets robotics: The KITTI dataset}, author={Geiger, Andreas and Lenz, Philip and Stiller, Christoph and Urtasun, Raquel}, journal={The International Journal of Robotics Research}, volume={32}, number={11}, pages={1231--1237}, year={2013}, publisher={SAGE Publications}}

@inproceedings{parkhi2012cats, title={Cats and dogs}, author={Parkhi, Omkar M and Vedaldi, Andrea and Zisserman, Andrew and Jawahar, C V}, booktitle={2012 IEEE Conference on Computer Vision and Pattern Recognition}, pages={3498--3505}, year={2012}, organization={IEEE}}

@article{cheng2017remote, title={Remote sensing image scene classification: Benchmark and state of the art}, author={Cheng, Gong and Han, Junwei and Lu, Xiaoqiang}, journal={Proceedings of the IEEE}, volume={105}, number={10}, pages={1865--1883}, year={2017}, publisher={IEEE}}

@inproceedings{everingham2010pascal, title={The Pascal visual object classes (VOC) challenge}, author={Everingham, Mark and Van Gool, Luc and Williams, Christopher KI and Winn, John and Zisserman, Andrew}, booktitle={International journal of computer vision}, volume={88}, pages={303--338}, year={2010}, organization={Springer}}

@inproceedings{joshi2024data,
  title={Data-efficient contrastive language-image pretraining: Prioritizing data quality over quantity},
  author={Joshi, Siddharth and Jain, Arnav and Payani, Ali and Mirzasoleiman, Baharan},
  booktitle={International Conference on Artificial Intelligence and Statistics},
  pages={1000--1008},
  year={2024},
  organization={PMLR}
}

@article{wu2023towards,
  title={Towards reporting bias in visual-language datasets: bimodal augmentation by decoupling object-attribute association},
  author={Wu, Qiyu and Zhao, Mengjie and He, Yutong and Huang, Lang and Ono, Junya and Wakaki, Hiromi and Mitsufuji, Yuki},
  journal={arXiv preprint arXiv:2310.01330},
  year={2023}
}

@inproceedings{yang2024clip,
  title={Clip-kd: An empirical study of clip model distillation},
  author={Yang, Chuanguang and An, Zhulin and Huang, Libo and Bi, Junyu and Yu, Xinqiang and Yang, Han and Diao, Boyu and Xu, Yongjun},
  booktitle={Proceedings of the IEEE/CVF Conference on Computer Vision and Pattern Recognition},
  pages={15952--15962},
  year={2024}
}

@article{chen2024comkd,
  title={Comkd-clip: Comprehensive knowledge distillation for contrastive language-image pre-traning model},
  author={Chen, Yifan and Qiao, Xiaozhen and Sun, Zhe and Li, Xuelong},
  journal={arXiv preprint arXiv:2408.04145},
  year={2024}
}

@inproceedings{chenglin2024mixed,
  title={Mixed distillation helps smaller language models reason better},
  author={Chenglin, Li and Chen, Qianglong and Li, Liangyue and Wang, Caiyu and Tao, Feng and Li, Yicheng and Chen, Zulong and Zhang, Yin},
  booktitle={Findings of the Association for Computational Linguistics: EMNLP 2024},
  pages={1673--1690},
  year={2024}
}

@article{xu2024survey,
  title={A survey on knowledge distillation of large language models},
  author={Xu, Xiaohan and Li, Ming and Tao, Chongyang and Shen, Tao and Cheng, Reynold and Li, Jinyang and Xu, Can and Tao, Dacheng and Zhou, Tianyi},
  journal={arXiv preprint arXiv:2402.13116},
  year={2024}
}

@inproceedings{cai2025llava,
  title={Llava-kd: A framework of distilling multimodal large language models},
  author={Cai, Yuxuan and Zhang, Jiangning and He, Haoyang and He, Xinwei and Tong, Ao and Gan, Zhenye and Wang, Chengjie and Xue, Zhucun and Liu, Yong and Bai, Xiang},
  booktitle={Proceedings of the IEEE/CVF International Conference on Computer Vision},
  pages={239--249},
  year={2025}
}

@article{xu2024llavadi,
  title={Llavadi: What matters for multimodal large language models distillation},
  author={Xu, Shilin and Li, Xiangtai and Yuan, Haobo and Qi, Lu and Tong, Yunhai and Yang, Ming-Hsuan},
  journal={arXiv preprint arXiv:2407.19409},
  year={2024}
}

@inproceedings{kornblith2019similarity,
  title={Similarity of neural network representations revisited},
  author={Kornblith, Simon and Norouzi, Mohammad and Lee, Honglak and Hinton, Geoffrey},
  booktitle={International conference on machine learning},
  pages={3519--3529},
  year={2019},
  organization={PMlR}
}

@article{dumpala2024sugarcrepe++,
  title={Sugarcrepe++ dataset: Vision-language model sensitivity to semantic and lexical alterations},
  author={Dumpala, Sri Harsha and Jaiswal, Aman and Shama Sastry, Chandramouli and Milios, Evangelos and Oore, Sageev and Sajjad, Hassan},
  journal={Advances in Neural Information Processing Systems},
  volume={37},
  pages={17972--18018},
  year={2024}
}

@inproceedings{singh2025learning,
  title={Learning the Power of “No”: Foundation Models with Negations},
  author={Singh, Jaisidh and Shrivastava, Ishaan and Vatsa, Mayank and Singh, Richa and Bharati, Aparna},
  booktitle={2025 IEEE/CVF Winter Conference on Applications of Computer Vision (WACV)},
  pages={8002--8012},
  year={2025},
  organization={IEEE}
}

@article{lavoie2024modeling,
  title={Modeling caption diversity in contrastive vision-language pretraining},
  author={Lavoie, Samuel and Kirichenko, Polina and Ibrahim, Mark and Assran, Mahmoud and Wilson, Andrew Gordon and Courville, Aaron and Ballas, Nicolas},
  journal={arXiv preprint arXiv:2405.00740},
  year={2024}
}

@inproceedings{zheng2024dreamlip,
  title={Dreamlip: Language-image pre-training with long captions},
  author={Zheng, Kecheng and Zhang, Yifei and Wu, Wei and Lu, Fan and Ma, Shuailei and Jin, Xin and Chen, Wei and Shen, Yujun},
  booktitle={European Conference on Computer Vision},
  pages={73--90},
  year={2024},
  organization={Springer}
}

@article{jiang2024vlm2vec,
  title={Vlm2vec: Training vision-language models for massive multimodal embedding tasks},
  author={Jiang, Ziyan and Meng, Rui and Yang, Xinyi and Yavuz, Semih and Zhou, Yingbo and Chen, Wenhu},
  journal={arXiv preprint arXiv:2410.05160},
  year={2024}
}

@article{meng2025vlm2vec,
  title={Vlm2vec-v2: Advancing multimodal embedding for videos, images, and visual documents},
  author={Meng, Rui and Jiang, Ziyan and Liu, Ye and Su, Mingyi and Yang, Xinyi and Fu, Yuepeng and Qin, Can and Chen, Zeyuan and Xu, Ran and Xiong, Caiming and others},
  journal={arXiv preprint arXiv:2507.04590},
  year={2025}
}

@article{bai2025qwen3,
  title={Qwen3-vl technical report},
  author={Bai, Shuai and Cai, Yuxuan and Chen, Ruizhe and Chen, Keqin and Chen, Xionghui and Cheng, Zesen and Deng, Lianghao and Ding, Wei and Gao, Chang and Ge, Chunjiang and others},
  journal={arXiv preprint arXiv:2511.21631},
  year={2025}
}

@misc{qwen3.5,
    title  = {{Qwen3.5}: Towards Native Multimodal Agents},
    author = {{Qwen Team}},
    month  = {February},
    year   = {2026},
    url    = {https://qwen.ai/blog?id=qwen3.5}
}

@article{wang2025internvl3,
  title={Internvl3. 5: Advancing open-source multimodal models in versatility, reasoning, and efficiency},
  author={Wang, Weiyun and Gao, Zhangwei and Gu, Lixin and Pu, Hengjun and Cui, Long and Wei, Xingguang and Liu, Zhaoyang and Jing, Linglin and Ye, Shenglong and Shao, Jie and others},
  journal={arXiv preprint arXiv:2508.18265},
  year={2025}
}

@article{qwen3vlembedding,
  title={Qwen3-VL-Embedding and Qwen3-VL-Reranker: A Unified Framework for State-of-the-Art Multimodal Retrieval and Ranking},
  author={Li, Mingxin and Zhang, Yanzhao and Long, Dingkun and Chen Keqin and Song, Sibo and Bai, Shuai and Yang, Zhibo and Xie, Pengjun and Yang, An and Liu, Dayiheng and Zhou, Jingren and Lin, Junyang},
  journal={arXiv preprint arXiv:2601.04720},
  year={2026}
}

@article{yang2025qwen3,
  title={Qwen3 technical report},
  author={Yang, An and Li, Anfeng and Yang, Baosong and Zhang, Beichen and Hui, Binyuan and Zheng, Bo and Yu, Bowen and Gao, Chang and Huang, Chengen and Lv, Chenxu and others},
  journal={arXiv preprint arXiv:2505.09388},
  year={2025}
}

@inproceedings{sauer2024adversarial,
  title={Adversarial diffusion distillation},
  author={Sauer, Axel and Lorenz, Dominik and Blattmann, Andreas and Rombach, Robin},
  booktitle={European Conference on Computer Vision},
  pages={87--103},
  year={2024},
  organization={Springer}
}

\appendix

\newpage

\setcounter{table}{0}
\renewcommand{\thetable}{\Alph{table}}

\setcounter{figure}{0}
\renewcommand{\thefigure}{\Alph{figure}}

\setcounter{equation}{0}
\renewcommand{\theequation}{\Alph{equation}}

\setcounter{section}{0}
\renewcommand{\thesection}{\Alph{section}}

\section{Additional Ablation Studies}\label{app:zs}
\paragraph{Data-level versus Feature-level Supervision.}
\begin{table}[t]
    \centering
    \caption{Data-level versus feature-level distillation with the same MLLM (Qwen3.5-2B) supplying both forms of supervision. The baseline rows are controlled reruns, not the main-table rows; TripletCLIP still renders its negative images with SDXL-Turbo.}
    \vspace{-0.25em}
    \small
    \resizebox{\linewidth}{!}{
        \begin{tabular}{ll|ccc}
            \toprule
            \textbf{Method} & \textbf{Supervision} & \textbf{Comp.} & \textbf{Zero-shot Cls.} & \textbf{Ret.} \\
            \midrule
            NegCLIP     & Data    & 37.9 & 25.0 & 12.9 \\
            TripletCLIP & Data    & 38.5 & 25.6 & 15.7 \\
            MLLMCLIP    & Feature & \textbf{40.0} & \textbf{31.3} & \textbf{19.1} \\
            \bottomrule
        \end{tabular}}
    \label{tab:4_abl_teacher}
\end{table}

In the main comparison, the data-level baselines and MLLMCLIP draw on different external models, so the gain could in principle be attributed to teacher identity rather than to feature-level transfer.
To rule this out, we control for teacher identity: the same MLLM (Qwen3.5-2B) serves both as the hard-negative generator for the data-level baselines and as the feature-level teacher for MLLMCLIP, under an identical training budget.
As shown in Table~\ref{tab:4_abl_teacher}, feature-level distillation still outperforms both data-level variants on all three axes once the teacher is controlled, with the widest margins on zero-shot classification and retrieval.
This comparison is also conservative with respect to the main tables, where the baselines rely on larger external models (LLaMA-7B, Qwen3-4B, and SDXL-Turbo) than our 2B teacher.

\paragraph{Teacher Prompt.}
\begin{table}[t]
    \centering
    \caption{Effect of the prompt used for teacher feature extraction. Both rows share the same Qwen3.5-2B teacher and training settings; only the extraction prompt differs.}
    \vspace{-0.25em}
    \small
    \resizebox{\linewidth}{!}{
        \begin{tabular}{l|ccc}
            \toprule
            \textbf{Teacher Prompt} & \textbf{Comp.} & \textbf{Zero-shot Cls.} & \textbf{Ret.} \\
            \midrule
            Plain caption (\texttt{\{caption\}} only) & 39.6 & 30.9 & 18.2 \\
            Compositional prompt (ours)    & \textbf{40.0} & \textbf{31.3} & \textbf{19.1} \\
            \bottomrule
        \end{tabular}}
    \label{tab:4_abl_prompt}
\end{table}

Teacher features are extracted with a fixed template that asks the MLLM to contrast the caption against a subtly incorrect alternative (Appendix~\ref{app:implementation}).
To attribute the gain to this compositional instruction rather than to generic conditioning on the caption, we re-extract teacher features with a plain caption prompt carrying no instruction, keeping the teacher and every other setting fixed.
Table~\ref{tab:4_abl_prompt} shows that the compositional prompt yields a small but consistent improvement on all three axes.
Notably, the plain-caption variant alone already surpasses the strongest data-level baseline of Table~\ref{tab:4_abl_teacher}, indicating that most of the benefit stems from transferring MLLM features at all, with the prompt acting as a further refinement.
A full prompt-by-prompt ablation would require re-extracting teacher features over the entire corpus for every variant, which is computationally prohibitive; we therefore report the single most informative comparison and leave broader prompt design to future work.

\paragraph{Teacher Layer Selection.}
\begin{table}[t]
    \centering
    \caption{Downstream performance with varying teacher layer selection strategies for distillation.}
    \vspace{-0.25em}
    \resizebox{\linewidth}{!}{
        \begin{tabular}{l|ccc}
            \toprule
            \textbf{Layer Selection} & \textbf{Comp.} & \textbf{Zero-shot Cls.} & \textbf{Ret.} \\
            \midrule
            Lower Block (1–12)           & 38.9 & 30.1 & 17.2 \\
            Middle Block (13–24)         & 38.2 & 28.6 & 16.3 \\
            Upper Block (25–36)          & 38.7 & 29.9 & 16.1 \\
            Strided (1,4,7...)           & \textbf{40.0} & \textbf{31.3} & \textbf{19.1} \\
            \bottomrule
        \end{tabular}
    }
    \label{tab:4_abl_layer}
\end{table}

We investigate the optimal strategy for selecting teacher layers for distillation, with results presented in Table~\ref{tab:4_abl_layer}.
We first evaluate strategies that use contiguous blocks of layers from the teacher MLLM: the lower (1–12), middle (13–24), and upper (25–36).
These block-based strategies yield comparable performance, with no clear advantage for any single block.
In contrast, a strided selection strategy that samples layers uniformly across the entire network (e.g., 1, 4, 7, ...) outperforms all block-based approaches.
This indicates that supervision derived from a broader range of layers provides more comprehensive guidance, whereas relying on a contiguous block of layers may overlook information distributed throughout the teacher model.

\paragraph{Weight Parameter.}
\begin{table}[t]
    \centering
    \small
    \caption{Downstream performance with varying weight parameter during pre-training.}
    \begin{tabular}{l|ccc}
            \toprule
            $\lambda$ & \textbf{Comp.} & \textbf{Zero-shot Cls.} & \textbf{Ret.} \\
            \midrule
            0.1      & 38.5 & 25.3 & 13.5 \\
            1        & 40.0 & \textbf{31.3} & \textbf{19.1} \\
            10       & \textbf{40.2} & 30.9 & 18.5 \\
            100      & 37.8 & 26.3 & 17.9 \\
            \bottomrule
        \end{tabular}
    \label{tab:4_abl_weight}
\end{table}

Table~\ref{tab:4_abl_weight} shows the impact of varying the distillation loss weight $\lambda$.
A small weight ($\lambda = 0.1$) provides insufficient supervision and a very large weight ($\lambda = 100$) over-constrains the student; both degrade performance across all metrics.
In the intermediate range, $\lambda = 1$ and $\lambda = 10$ perform comparably: $\lambda = 1$ achieves the best zero-shot classification and retrieval, while $\lambda = 10$ marginally improves compositionality.
We adopt $\lambda = 1$ as the default for a balanced trade-off across the three evaluation groups.

\section{Experimental Settings}
\subsection{Evaluation Benchmarks}\label{app:dataset}
\paragraph{Compositionality.}
We evaluate on 11 compositionality benchmarks: ARO~\citep{yuksekgonul2022and}, CREPE~\citep{ma2023crepe}, EQBEN~\citep{wang2023equivariant}, ImageCoDe~\citep{krojer2022image}, SugarCrepe~\citep{hsieh2023sugarcrepe}, SVO-Probes~\citep{hendricks2021probing}, VALSE~\citep{parcalabescu2021valse}, VL-Checklist~\citep{zhao2022vl}, WhatsUp~\citep{kamath2023s}, Winoground~\citep{thrush2022winoground}, and SPEC~\citep{peng2024synthesize}.

\paragraph{Retrieval.}
We use MSCOCO~\citep{chen2015microsoft} and Flickr-30K~\citep{plummer2015flickr30k} for zero-shot image-to-text and text-to-image retrieval.

\paragraph{Classification.}
We evaluate zero-shot classification on 13 datasets: Caltech101~\citep{fei2007learning}; CIFAR-10, CIFAR-100~\citep{krizhevsky2009learning}; Describable Textures~\citep{cimpoi14describing}; EuroSAT-CLIP~\cite{helber2019eurosat}; FER-2013~\citep{goodfellow2013challenges}; Flower102~\citep{Nilsback08}; Food101~\citep{bossard14}; ImageNet-1K~\citep{deng2009imagenet}; KITTI-Distance~\citep{geiger2013vision}; Oxford-IIIT Pet~\citep{parkhi2012cats}; RESISC45-CLIP~\citep{cheng2017remote}; and PASCAL VOC2007~\citep{everingham2010pascal}.

\begin{table*}[t!]
\centering
\small
\caption{Public sources of the models used in our experiments.}
\label{tab:model_sources}
\begin{tabular}{ll}
\toprule
Model & Source \\
\midrule
Qwen3-4B (hard-negative captions) & \url{https://huggingface.co/Qwen/Qwen3-4B} \\
SDXL-Turbo (hard-negative images) & \url{https://huggingface.co/stabilityai/sdxl-turbo} \\
\midrule
LLaVA-1.6-Mistral-7B & \url{https://huggingface.co/llava-hf/llava-v1.6-mistral-7b-hf} \\
LLaVA-1.6-Vicuna-7B & \url{https://huggingface.co/llava-hf/llava-v1.6-vicuna-7b-hf} \\
LLaMA-3.2-11B-Vision-Instruct & \url{https://huggingface.co/meta-llama/Llama-3.2-11B-Vision-Instruct} \\
Qwen2-VL-2B-Instruct & \url{https://huggingface.co/Qwen/Qwen2-VL-2B-Instruct} \\
Qwen3-VL-2B-Instruct & \url{https://huggingface.co/Qwen/Qwen3-VL-2B-Instruct} \\
Qwen3-VL-4B-Instruct & \url{https://huggingface.co/Qwen/Qwen3-VL-4B-Instruct} \\
Qwen3.5-2B & \url{https://huggingface.co/Qwen/Qwen3.5-2B} \\
Qwen3.5-4B & \url{https://huggingface.co/Qwen/Qwen3.5-4B} \\
InternVL3-2B & \url{https://huggingface.co/OpenGVLab/InternVL3-2B} \\
InternVL3-8B & \url{https://huggingface.co/OpenGVLab/InternVL3-8B} \\
InternVL3.5-2B & \url{https://huggingface.co/OpenGVLab/InternVL3_5-2B} \\
InternVL3.5-4B & \url{https://huggingface.co/OpenGVLab/InternVL3_5-4B} \\
\midrule
VLM2Vec-v1 (Qwen2-VL-2B) & \url{https://huggingface.co/TIGER-Lab/VLM2Vec-Qwen2VL-2B} \\
VLM2Vec-v1 (Qwen2-VL-7B) & \url{https://huggingface.co/TIGER-Lab/VLM2Vec-Qwen2VL-7B} \\
VLM2Vec-v1 (LLaVA-Next) & \url{https://huggingface.co/TIGER-Lab/VLM2Vec-LLaVa-Next} \\
VLM2Vec-v1 (Phi-3.5V) & \url{https://huggingface.co/TIGER-Lab/VLM2Vec-Full} \\
VLM2Vec-V2.0 & \url{https://huggingface.co/VLM2Vec/VLM2Vec-V2.0} \\
Qwen3-VL-Embedding-2B & \url{https://huggingface.co/Qwen/Qwen3-VL-Embedding-2B} \\
Qwen3-VL-Embedding-8B & \url{https://huggingface.co/Qwen/Qwen3-VL-Embedding-8B} \\
\bottomrule
\end{tabular}
\end{table*}

\subsection{Implementation Details}\label{app:implementation}
\paragraph{Teacher Feature-Extraction Prompt.}
The full prompt template is shown below.

\begin{promptbox}[title=Teacher prompt (feature extraction)]
\small
Given the image and the caption \texttt{\{caption\}}, analyze whether this caption accurately describes the image.
If it does, imagine a similar caption that could be easily confused with it but is subtly incorrect or misleading.
Internally reason about the difference between the correct and incorrect caption, highlighting the key visual–semantic concepts that make the original caption more accurate.
Focus on compositional elements such as object attributes, actions, relationships, and spatial arrangements.
Use this reasoning to build an internal representation of the image that emphasizes these distinctions.
\end{promptbox}

\paragraph{Training Hyperparameters.}
We train all models on 8 NVIDIA A100 GPUs using the AdamW optimizer with a batch size of 4096, an initial learning rate of $5 \times 10^{-4}$, and a weight decay of 0.5 for 30 epochs.
A cosine learning rate scheduler is applied with a linear warmup during the first epoch, where the learning rate increases from $1 \times 10^{-6}$ to the base learning rate and decays to $1 \times 10^{-5}$ by the end of training.
All experiments are conducted using \texttt{bfloat16} precision. Unless otherwise specified, we use a fixed loss weight of $\lambda = 1$ for balancing the distillation and contrastive losses.

\subsection{Model Sources}\label{app:models}
Table~\ref{tab:model_sources} lists the public sources of all models used in our experiments.

\section*{Reproducibility Statement}
We provide all implementation details, including model architectures, training hyperparameters, and evaluation protocols, in the main paper and appendix. All experiments are conducted using publicly available datasets, and we will release our code, pretrained models, and data-processing scripts to ensure reproducibility.

\section*{LLM Usage}
During the preparation of this paper, we use large language models in a limited and assistive manner.
For implementation, an LLM is utilized for code review and the detection of minor bugs.
For writing, an LLM is used for English proofreading and grammar checks.
We do not have LLMs draft whole passages or sentences from scratch, nor do we rely on them to generate novel methods or results.

\begin{table*}[t]
\centering
\small
\caption{Comparison of MLLM-as-judge models and MLLM-based embedding models on the SugarCrepe benchmark. Judge models are queried twice per example with the ground-truth caption at position A (\emph{First}) and position B (\emph{Second}); embedding models score each caption by image–text similarity and report a single per-task accuracy.}
\resizebox{\textwidth}{!}{%
\begin{tabular}{llcccccccc}
\toprule
Model & GT-position / Backbone
  & \multicolumn{3}{c}{Replace}
  & \multicolumn{2}{c}{Swap}
  & \multicolumn{2}{c}{Add}
  & \multirow{2}{*}{Average}  \\
\cmidrule(lr){3-5} \cmidrule(lr){6-7} \cmidrule(lr){8-9}
      &
  & Object & Attribute & Relation
  & Object & Attribute
  & Object & Attribute
  &  \\
\midrule
\multicolumn{10}{l}{\textit{MLLM-as-Judge models}} \\
\midrule
\multirow{3}{*}{LLAVA-1.6-mistral-7B}
  & First   & 99.3 & 98.0 & 95.8 & 89.4 & 96.2 & 97.2 & 93.1 & 95.6 \\
  & Second  & 93.8 & 82.6 & 67.6 & 69.4 & 77.6 & 79.3 & 50.0 & 74.3 \\
  & Average & 96.5 & 90.3 & 81.7 & 79.4 & 86.9 & 88.2 & 71.5 & 85.0 \\
\midrule
\multirow{3}{*}{LLAVA-1.6-vicuna-7B}
  & First   & 98.1 & 96.1 & 89.6 & 80.8 & 89.9 & 93.9 & 85.5 & 90.6 \\
  & Second  & 93.6 & 75.6 & 82.0 & 78.4 & 74.8 & 83.4 & 68.9 & 79.5 \\
  & Average & 95.9 & 85.9 & 85.8 & 79.6 & 82.4 & 88.7 & 77.2 & 85.1 \\
\midrule
\multirow{3}{*}{LLAMA-3.2-Vision-11B}
  & First   & 98.4 & 95.9 & 89.5 & 85.3 & 95.5 & 95.6 & 84.5 & 92.1 \\
  & Second  & 97.8 & 92.5 & 90.5 & 90.6 & 91.6 & 93.6 & 90.6 & 92.5 \\
  & Average & 98.1 & 94.2 & 90.0 & 88.0 & 93.5 & 94.6 & 87.6 & 92.3 \\
\midrule
\multirow{3}{*}{Qwen2-VL-2B}
  & First   & 96.7 & 91.4 & 87.1 & 76.3 & 91.3 & 95.5 & 84.0 & 88.9 \\
  & Second  & 98.5 & 96.3 & 93.2 & 92.2 & 95.2 & 97.3 & 95.2 & 95.4 \\
  & Average & 97.6 & 93.9 & 90.2 & 84.3 & 93.2 & 96.4 & 89.6 & 92.2 \\
\midrule
\multirow{3}{*}{Qwen3-VL-2B}
  & First   & 98.5 & 95.8 & 92.0 & 87.8 & 94.9 & 97.4 & 91.6 & 94.0 \\
  & Second  & 98.6 & 96.1 & 91.0 & 91.8 & 96.5 & 97.5 & 93.6 & 95.0 \\
  & Average & 98.6 & 95.9 & 91.5 & 89.8 & 95.7 & 97.4 & 92.6 & 94.5 \\
\midrule
\multirow{3}{*}{Qwen3-VL-4B}
  & First   & 98.6 & 95.8 & 93.2 & 94.3 & 96.7 & 97.5 & 94.4 & 95.8 \\
  & Second  & 99.0 & 97.2 & 94.2 & 92.2 & 97.9 & 97.9 & 96.5 & 96.4 \\
  & Average & 98.8 & 96.5 & 93.7 & 93.3 & 97.3 & 97.7 & 95.4 & 96.1 \\
\midrule
\multirow{3}{*}{Qwen3.5-2B}
  & First   & 97.0 & 92.8 & 88.5 & 80.4 & 94.0 & 88.5 & 78.5 & 88.5 \\
  & Second  & 99.0 & 98.4 & 93.2 & 94.7 & 96.7 & 97.2 & 93.1 & 96.0 \\
  & Average & 98.0 & 95.6 & 90.9 & 87.6 & 95.3 & 92.8 & 85.8 & 92.3 \\
\midrule
\multirow{3}{*}{Qwen3.5-4B}
  & First   & 97.5 & 93.5 & 91.5 & 89.0 & 95.0 & 94.9 & 81.6 & 91.9 \\
  & Second  & 99.5 & 98.6 & 97.9 & 94.3 & 99.7 & 99.0 & 99.0 & 98.3 \\
  & Average & 98.5 & 96.1 & 94.7 & 91.6 & 97.4 & 96.9 & 90.3 & 95.1 \\
\midrule
\multirow{3}{*}{InternVL3-2B}
  & First   & 98.8 & 96.2 & 94.2 & 91.0 & 96.5 & 97.5 & 91.2 & 95.1 \\
  & Second  & 98.0 & 93.1 & 89.0 & 83.3 & 92.0 & 95.0 & 88.6 & 91.3 \\
  & Average & 98.4 & 94.7 & 91.6 & 87.1 & 94.3 & 96.2 & 89.9 & 93.2 \\
\midrule
\multirow{3}{*}{InternVL3-8B}
  & First   & 98.6 & 96.4 & 93.0 & 95.1 & 97.4 & 96.6 & 87.9 & 95.0 \\
  & Second  & 98.7 & 97.1 & 94.2 & 93.1 & 98.0 & 96.3 & 96.1 & 96.2 \\
  & Average & 98.7 & 96.8 & 93.6 & 94.1 & 97.7 & 96.4 & 92.0 & 95.6 \\
\midrule
\multirow{3}{*}{InternVL3.5-2B}
  & First   & 98.1 & 95.3 & 93.9 & 90.2 & 94.6 & 95.5 & 92.1 & 94.2 \\
  & Second  & 96.7 & 90.6 & 84.8 & 85.3 & 89.6 & 91.7 & 87.1 & 89.4 \\
  & Average & 97.4 & 93.0 & 89.3 & 87.8 & 92.1 & 93.6 & 89.6 & 91.8 \\
\midrule
\multirow{3}{*}{InternVL3.5-4B}
  & First   & 97.6 & 94.4 & 93.1 & 91.8 & 97.4 & 95.1 & 89.3 & 94.1 \\
  & Second  & 97.9 & 94.3 & 92.3 & 87.3 & 96.1 & 94.0 & 92.8 & 93.5 \\
  & Average & 97.8 & 94.4 & 92.7 & 89.6 & 96.8 & 94.6 & 91.0 & 93.8 \\
\midrule
\multicolumn{10}{l}{\textit{Embedding models}} \\
\midrule
VLM2Vec-v1  & LLAVA-1.6-mistral-7B & 81.7 & 71.7 & 61.5 & 51.6 & 56.6 & 74.5 & 69.6 & 66.7 \\
VLM2Vec-v1  & phi3.5V     & 83.5 & 72.3 & 60.2 & 53.9 & 57.4 & 73.9 & 69.4 & 67.2 \\
VLM2Vec-v1  & Qwen2-VL-2B & 89.5 & 73.1 & 61.5 & 42.0 & 49.5 & 72.9 & 63.9 & 64.6 \\
VLM2Vec-v1  & Qwen2-VL-7B & 86.4 & 72.2 & 61.7 & 41.6 & 53.2 & 76.1 & 71.1 & 66.0 \\
VLM2Vec-v2  & Qwen2-VL-2B & 94.3 & 83.6 & 73.3 & 53.9 & 62.5 & 79.4 & 62.0 & 72.7 \\
Qwen3-VL-Embedding & Qwen3-VL-2B & 97.1 & 89.3 & 76.7 & 70.2 & 75.1 & 89.7 & 82.5 & 83.0 \\
Qwen3-VL-Embedding & Qwen3-VL-8B & 97.9 & 91.2 & 81.4 & 69.4 & 84.5 & 93.1 & 84.1 & 86.0 \\
\bottomrule
\end{tabular}%
}
\label{tab:full_sugarcrepe_results}
\end{table*}

\end{document}